\documentclass[10pt,twocolumn,letterpaper]{article}

\PassOptionsToPackage{table}{xcolor}
\usepackage{preprint}
\usepackage{microtype}

\makeatletter
\renewcommand{\paragraph}{%
  \@startsection{paragraph}{4}{\z@}%
  {.5em \@plus .1em \@minus .05em}%
  {-0.6em}%
  {\normalfont\normalsize\bfseries}%
}
\makeatother

\definecolor{figblue}{HTML}{174EA6}
\definecolor{figgreen}{HTML}{188038}
\definecolor{figred}{HTML}{C5221F}
\definecolor{figamber}{HTML}{B06000}

\usepackage{amsmath}
\usepackage{amssymb}
\usepackage{mathtools}
\usepackage{amsthm}

\theoremstyle{plain}

\theoremstyle{definition}

\theoremstyle{remark}

\definecolor{linkblue}{rgb}{0.21,0.49,0.74}
\usepackage[pagebackref,breaklinks,colorlinks,allcolors=linkblue]{hyperref}

\title{Video-as-Weights: Compiling a Video Once into Parametric and Token Memory}

\author{%
Wentao Zhang$^{1,*}$ \quad Ian Miao$^{2}$ \quad Chen Qu$^{2}$ \quad Bill Bi$^{2}$ \\
Donna Schut$^{2}$ \quad Yuntian Deng$^{1}$ \quad Yale Song$^{2}$ \\[2pt]
{\normalsize $^{1}$University of Waterloo \qquad $^{2}$Google} \\
{\tt\small \{w564zhan,yuntian\}@uwaterloo.ca \qquad \{yilinmiao,cqu,chujianb,dschut,yalesong\}@google.com}%
}

\makeatletter
\def\@maketitle{%
  \newpage
  \null
  \vskip .15in
  \begin{center}%
    {\Large \bf \@title \par}%
    \vspace*{12pt}%
    {\large
      \lineskip .5em
      \begin{tabular}[t]{c}%
        \@author
      \end{tabular}\par}%
    \vskip .5em
  \end{center}%
}
\makeatother

\begin{document}
\twocolumn[{%
\renewcommand\twocolumn[1][]{#1}%
\maketitle
{
  \centering
  \expandafter\def\csname @captype\endcsname{figure}
  \includegraphics[height=0.3\textwidth]{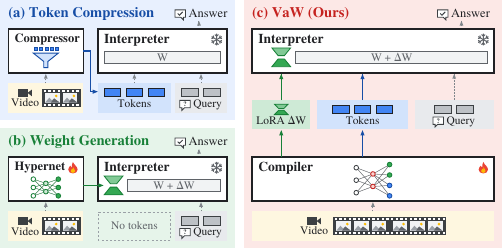}%
  \hfill
  \includegraphics{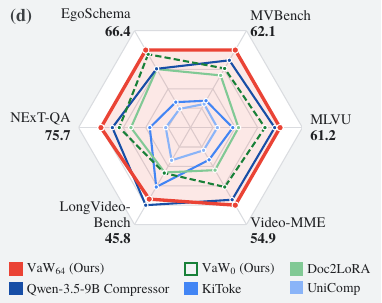}
  \caption{\textbf{Video as Weights (VaW) vs.\ existing approaches.}
  \textcolor{figblue}{\textbf{(a) Token compression}} feeds the video as a few compressed or pruned visual tokens, e.g., a 9B~$\rightarrow$~0.8B compressor, KiToke \cite{huang2026kitokekernelbasedintervalawaretoken}, and UniComp \cite{yuan2026unicomp}.
  \textcolor{figgreen}{\textbf{(b) Weight generation}} turns the video into weights, with no tokens, e.g., Doc2LoRA \cite{charakorn2026doctolora}.
  \textcolor{figred}{\textbf{(c) VaW (ours)}} compiles the video into both LoRA weights and optional tokens.
  \textbf{(d)} VaW outperforms both families on five of six video benchmarks and on average. Subscripts denote the number of visual tokens.}
  \label{fig:single}
  \par
}
\vskip 0.02in

}]%
\begingroup
\renewcommand\thefootnote{$*$}%
\begin{NoHyper}\footnotetext{Work done during Wentao Zhang's internship at Google.}\end{NoHyper}%
\endgroup%
\begin{abstract}
Vision-language models (VLMs) represent videos as thousands of visual tokens, creating a large key-value cache whose memory grows with video length. At the same time, small VLMs that fit on limited hardware understand video far worse than large models. Existing compression methods shorten visual tokens into a brief prompt to save memory, yet the small model still processes them through fixed weights and cannot close this capability gap.

To address both challenges, we introduce Video-as-Weights (VaW), which compiles a video once without seeing any question into the parametric and token memory of a frozen interpreter VLM. Specifically, a compiler VLM watches the video, and a single weight decoder jointly generates the interpreter's LoRA weights and optional prefix tokens from every compiler layer, enabling both same-model compilation and strong-to-weak transfer across model sizes and depths.

Across six video QA benchmarks, a 0.8B interpreter beats full in-context reading by \textbf{1.7 points} when compiled by itself and by \textbf{4.3 points} with a 9B compiler, reducing context from up to \textbf{8,192} visual tokens to \textbf{64} prefix tokens while using \textbf{5.5$\times$} less memory than the video's key-value cache. Generated weights add \textbf{3.9 points} over token compression alone, and all-layer decoding outperforms layer-wise generation by \textbf{2.9 points}. Mechanistic analysis shows that with prefix tokens, the weights mainly alter attention queries and MLP down projections to guide retrieval and store a baseline summary, whereas without tokens, they shift to attention output projections to write video content directly into the residual stream.
\end{abstract}

\section{Introduction}

Video understanding in practice is asymmetric: a video is ingested once, then queried many times across turns and questions.
Standard vision-language models (VLMs) do not exploit this, keeping the video in context as thousands of visual tokens whose key-value (KV) cache grows with video length.
Deploying a small VLM cuts the model size, yet small models are much weaker at video understanding than large ones.

Because a video is ingested only once, we can compress it into a compact memory and, when a stronger model is available, transfer that model's understanding to the answering model.

Most prior work cuts this cost in \emph{token space} (\Cref{fig:single}a), either by training models to pack frame features into fewer tokens \cite{NEURIPS2022_960a172b,llavamini,shen2024longvu,li2026videochatflash} or by pruning and merging the visual tokens of an existing model \cite{Yang_2025_CVPR,shen2025fastvid,huang2026kitokekernelbasedintervalawaretoken,yuan2026unicomp}.
Compression that ignores the question cannot tell which details will matter, so heavy pruning drops fine details and hurts accuracy, while compression conditioned on the question \cite{fei2026small} must be run again for every new question.
Either way, the small model still reads these tokens with its own fixed weights, so the capacity gap remains.

Another approach moves context into \emph{parametric space} (\Cref{fig:single}b).
For instructions, prior work uses hypernetworks \cite{ha2017hypernetworks} to compile a prompt once into adapters and prefixes instead of reading it again for each input \cite{ivison-etal-2023-hint,3618408.3619566,charakorn2025texttolora}, while Program-as-Weights \cite{zhang2026programasweightsprogrammingparadigmfuzzy} lets a larger compiler write task programs for a smaller interpreter.
For content, text documents \cite{chen2025generative,charakorn2026doctolora,liu2026shine} and visual frames \cite{ma2024visual,suri2026frames2loraparametricvideointernalization} have been written into a model's own weights.
A video is much harder than an instruction, because it holds far more content and different questions depend on different parts of it.
Existing generators either tie each generated layer to one source layer \cite{charakorn2026doctolora,suri2026frames2loraparametricvideointernalization} or pool the source into a single vector \cite{zhang2026programasweightsprogrammingparadigmfuzzy}.
Also, methods that store content in weights use weights alone, without tokens to carry fine details.

\begin{figure*}[t]
  \centering
  \includegraphics[width=\textwidth]{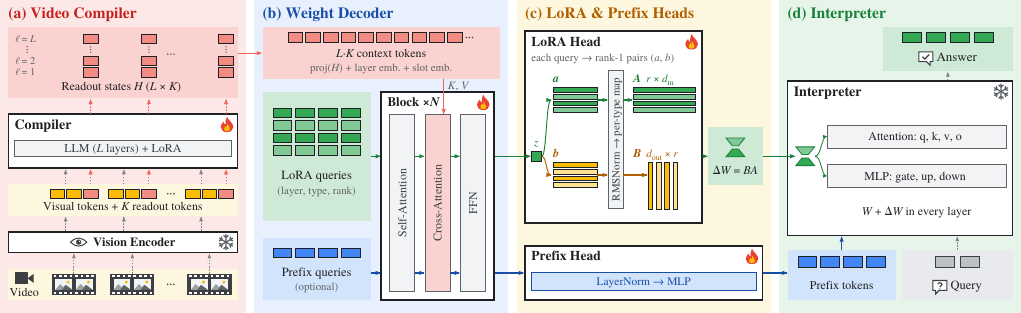}
  \caption{\textbf{VaW architecture.}
  \textcolor{figred}{\textbf{(a) Video compiler:}} a frozen vision encoder and LoRA-adapted LLM read the video with $K$ readout tokens, outputting readout states $H$ across all $L$ layers.
  \textcolor{figblue}{\textbf{(b) Weight decoder:}} $N$ blocks refine learned LoRA and optional prefix queries by cross-attending to $H$.
  \textcolor{figamber}{\textbf{(c) LoRA \& prefix heads:}} decoder states map to rank-$r$ updates $\Delta W = BA$ and optional prefix tokens.
  \textcolor{figgreen}{\textbf{(d) Interpreter:}} the frozen model applies $\Delta W$ across layers and reads any prefix tokens before the query. Panels (a)--(c) never see the query, so compilation runs once per video; flame marks trained parts and snowflake marks frozen parts.}
  \label{fig:arch}
\end{figure*}

To solve both problems, we introduce \textbf{Video-as-Weights (VaW)}, which treats a video as a program to compile rather than a document to read (\Cref{fig:single}c and \Cref{fig:arch}).
A \emph{video compiler} VLM watches the video once, without seeing the question, and a \emph{weight decoder} turns its hidden states into a compact parametric and token memory (LoRA weights and a few prefix tokens) for a frozen \emph{interpreter} VLM.
The interpreter then answers every question without reading any visual tokens.
The weights change how the interpreter processes each question, not just what it reads, and because they are decoded jointly with the tokens, the two channels split the video's content between them.
Every generated weight can draw on any compiler layer, which we call \emph{all-layer decoding}, so the compiler can match the interpreter or be a larger, deeper model.

On six video question answering benchmarks, VaW lets a 0.8B interpreter beat full in-context reading by 1.7 points with a 0.8B compiler and by 4.3 points with a 9B compiler, using 64 tokens instead of up to 8,192 and $5.5\times$ less memory than the video's KV cache.
Our contributions are as follows:
\begin{itemize}
    \setlength{\itemsep}{2pt}
    \setlength{\topsep}{2pt}
    \setlength{\partopsep}{0pt}
    \item \textbf{Video-as-Weights.} We introduce a compile-once framework that writes a video into both the LoRA weights (parametric memory) and prefix tokens (token memory) of a frozen interpreter VLM. Both channels help: the tokens add about 5 points over the weights alone, and the weights add 5.1 points over the tokens alone.
    \item \textbf{Strong-to-weak transfer through all-layer decoding.} All-layer decoding beats a layer-wise decoder by 2.9 points and lets a 9B compiler write for a 0.8B interpreter of a different depth, improving accuracy by 2.6 points over a 0.8B compiler and by 3.9 points over token compression with the same 9B compiler.
    \item \textbf{Compact per-video memory.} VaW packs up to 8,192 visual tokens into a fixed 20.6\,MB state ($5.5\times$ smaller than the interpreter's KV cache), beating full in-context reading by 1.7 and 4.3 points (with 0.8B and 9B compilers) and coming within 1.2 points of full-video SFT.
    \item \textbf{Analysis.} We show how the two channels split the work: with tokens, the generated weights mainly guide the interpreter's attention, and without tokens, they carry the video content directly.
\end{itemize}

\section{Related Work}

\paragraph{Token-efficient VLMs and visual token pruning.}
Most work on long videos reduces the number of visual tokens a VLM reads.
Some models are built with a bottleneck that represents frames with fewer tokens, such as a resampler \cite{NEURIPS2022_960a172b}, one token per frame \cite{llavamini}, or adaptive and hierarchical compression \cite{shen2024longvu,li2026videochatflash}.
Other methods prune or merge the visual tokens of a pretrained VLM, often without training \cite{Yang_2025_CVPR,shen2025fastvid,huang2026kitokekernelbasedintervalawaretoken,yuan2026unicomp}.
Compression that ignores the question can drop the details a later question needs, and Tempo \cite{fei2026small}, which uses a small VLM to compress the video for each question, must run again for every new question.
In both cases the small model reads the compressed tokens with its own weights, so its capacity stays the same.
VaW also gives the interpreter only 64 tokens, but a larger compiler writes them once, without the question, together with generated weights that change how the interpreter reads them.
These weights add 5.1 points over a tokens-only model trained the same way (\Cref{tab:ablation}).

\paragraph{Hypernetworks and parameter generation.}
Another line moves a context from the prompt into the weights.
Context distillation \cite{snell2023learning} does this with gradient updates for each context, while hypernetworks \cite{ha2017hypernetworks} generate the weights in one forward pass.
For instructions, HINT \cite{ivison-etal-2023-hint} and HyperTuning \cite{3618408.3619566} turn an instruction or a few examples into adapters or prefixes, so they need not be reread for every input, and Text-to-LoRA \cite{charakorn2025texttolora} turns a task description into a LoRA.
Program-as-Weights \cite{zhang2026programasweightsprogrammingparadigmfuzzy} uses a 4B compiler to write a LoRA for a 0.6B interpreter from a function description, and it pools the compiler's hidden states into a single vector before generating the weights.
For content, Generative Adapter \cite{chen2025generative}, Doc-to-LoRA \cite{charakorn2026doctolora}, and SHINE \cite{liu2026shine} write a text document into the adapter weights of the same model.
VLoRA \cite{ma2024visual} and Frames2LoRA \cite{suri2026frames2loraparametricvideointernalization} turn images and video frames into LoRA weights, and HyperLLaVA \cite{zhang2024hyperllava} generates visual and language experts while still reading every visual token.
VaW differs from these methods in three ways.
First, its compiler can be a separate, larger VLM. Methods that write content do not use a stronger model as the writer, and Program-as-Weights, which does, compiles a short task description rather than a long video.
Second, every generated weight in VaW can read every compiler layer at every readout position. Doc-to-LoRA and Frames2LoRA instead generate each layer from the matching source layer, and Program-as-Weights reduces the compiler to one vector. All-layer decoding outperforms a layer-wise decoder by 2.9 points (\Cref{tab:main_results}).
Third, VaW decodes LoRA weights and prefix tokens jointly. Methods that write content use weights alone or keep every visual token, and Program-as-Weights found that adding a prefix did not help its task programs. For video, the tokens add about 5 points over the weights alone.

\paragraph{Test-time adaptation and dynamic memory.}
A third line gives a model a memory that changes at inference.
Fast-weight language models \cite{clark-etal-2022-meta} update some layers from the preceding context, test-time training layers \cite{pmlr-v267-sun25h} use a hidden state that is itself a model trained on the test sequence, and Titans \cite{behrouz2025titans} learns a long-term memory that memorizes at test time.
These memories are part of the architecture, so the model must be trained with them, and the model writes its own memory.
Cache-to-Cache \cite{fu2026cachetocache} passes information from one LLM to another by projecting the KV cache of one into that of the other, but both models still read the full input, and the fused cache is as long as the input.
VaW instead gives a frozen VLM a fixed-sized memory, set in one forward pass by a different model, with no gradient steps at inference.

\section{Video-as-Weights}

\subsection{Model Architecture}
\label{sec:arch}

\paragraph{Overview.}
Given a video $V$ and a text query $x$, a standard VLM produces the answer $y$ from $p(y \mid V, x)$ by reading every visual token of $V$ in its context.
As shown in \Cref{fig:arch}, VaW instead compiles the video into the parametric and token memory of a frozen \emph{interpreter} VLM with parameters $W$ and hidden size $d$.
Inspired by Program-as-Weights \cite{zhang2026programasweightsprogrammingparadigmfuzzy}, a \emph{video compiler} $\mathcal{C}$ reads the video, and a \emph{weight decoder} $\mathcal{D}$ turns what it read into a low-rank weight update $\Delta W$ and, optionally, $P$ prefix tokens $\mathbf{P} \in \mathbb{R}^{P \times d}$:
\begin{equation}
  (\Delta W, \mathbf{P}) = \mathcal{D}\big(\mathcal{C}(V)\big), \qquad
  y \sim p_{W + \Delta W}(y \mid \mathbf{P}, x).
  \label{eq:overview}
\end{equation}
The interpreter never sees a visual token: the video reaches it only through $\Delta W$ and $\mathbf{P}$.
Neither $\mathcal{C}$ nor $\mathcal{D}$ sees the query, so each video is compiled once and its $(\Delta W, \mathbf{P})$ is reused for every question about it.
The compiler and the interpreter are separate pretrained VLMs and may differ in size.
The prefix tokens are optional at inference, so one trained model can run with or without $\mathbf{P}$, as detailed in \Cref{sec:objective}.

\paragraph{Video compiler.}
The compiler, illustrated in \Cref{fig:arch}a, is a pretrained VLM with a vision encoder and an $L$-layer LLM of hidden size $d_c$.
The frozen vision encoder turns the sampled frames into visual tokens, grouped into temporal segments of consecutive frames.
We add $K$ \emph{readout tokens} to this sequence: they are divided evenly among the segments, and each share is placed right after the visual tokens of its segment.
$K$ grows with the number of visual tokens $n_v$ as $K \approx n_v / \rho$, rounded to a power of two and clipped to $[K_{\min}, K_{\max}]$, where $\rho$ is the readout ratio.
Readout tokens use learned embeddings indexed by their relative position $k/K$, so slot $k$ marks the same fraction of the video for any $K$.

The LLM, adapted with trainable LoRA \cite{hu2022lora} of rank $r_c$, reads this sequence causally.
Readout token $k$ thus summarizes the video up to its segment, and the last one summarizes the whole video.
We collect the hidden states at the readout positions from all layers,
\begin{equation}
  H = \big[\mathbf{h}_{\ell,k}\big] \in \mathbb{R}^{L \times K \times d_c},
\end{equation}
where $\mathbf{h}_{\ell,k}$ is the output of layer $\ell \in \{1, \dots, L\}$ at readout token $k \in \{1, \dots, K\}$.
Reading every layer, not only the last, gives the decoder both shallow and deep video features.

\paragraph{Weight decoder.}
\emph{Context.}
In the weight decoder of \Cref{fig:arch}b, each readout state becomes one context token of width $d_h$:
\begin{equation}
  \mathbf{c}_{\ell,k} = \operatorname{LN}\big(W_{\mathrm{in}} \mathbf{h}_{\ell,k} + \mathbf{e}^{\mathrm{layer}}_{\ell} + W_{\mathrm{pos}} \phi(k/K)\big),
\end{equation}
where $W_{\mathrm{in}} \in \mathbb{R}^{d_h \times d_c}$ is a projection, $\mathbf{e}^{\mathrm{layer}}_{\ell}$ is a learned layer embedding, and the slot embedding $W_{\mathrm{pos}} \phi(k/K)$ is a linear map of Fourier features $\phi$ of the relative position $k/K$.
The context $\mathbf{C} = \{\mathbf{c}_{\ell,k}\}$ has $L \cdot K$ tokens.

\emph{Queries.}
We adapt a set $\mathcal{S}$ of linear modules (\emph{sites}) in the interpreter.
A site $s = (j, \tau)$ is the module of type $\tau$ (e.g., the query projection or the MLP up projection) in interpreter layer $j$, with weight $W_s \in \mathbb{R}^{d^{\tau}_{\mathrm{out}} \times d^{\tau}_{\mathrm{in}}}$.
Its rank-$r$ update is a sum of $r$ rank-1 components, which we split into $m = r/c$ \emph{rank groups} of $c$ components each.
Each (site, rank group) pair has one learned query,
\begin{equation}
  \mathbf{q}_{s,g} = \mathbf{e}^{\mathrm{site}}_{s} + \mathbf{e}^{\mathrm{group}}_{g}, \qquad s \in \mathcal{S},\; g = 1, \dots, m,
\end{equation}
giving $|\mathcal{S}| \cdot m$ LoRA queries.
With prefix tokens enabled, we add $P$ learned prefix queries $\mathbf{q}^{\mathrm{pre}}_{1}, \dots, \mathbf{q}^{\mathrm{pre}}_{P}$.

\emph{Blocks.}
All queries, stacked as $\mathbf{Z}$, pass through $N$ pre-norm blocks:
\begin{align}
  \mathbf{Z} &\leftarrow \mathbf{Z} + \operatorname{SelfAttn}\big(\operatorname{LN}(\mathbf{Z})\big), \nonumber \\
  \mathbf{Z} &\leftarrow \mathbf{Z} + \operatorname{CrossAttn}\big(\operatorname{LN}(\mathbf{Z}), \mathbf{C}\big), \\
  \mathbf{Z} &\leftarrow \mathbf{Z} + \operatorname{FFN}\big(\operatorname{LN}(\mathbf{Z})\big), \nonumber
\end{align}
where cross-attention takes its keys and values from $\mathbf{C}$ and the FFN is a SwiGLU MLP.
Cross-attention lets every query read any compiler layer at any point in time.
Self-attention lets the queries coordinate, so that different rank groups, sites and prefix queries can divide the video's content among themselves.
We write $\mathbf{z}_{s,g}$ and $\mathbf{z}^{\mathrm{pre}}_{p}$ for the final query states.

\paragraph{LoRA head.}
In \Cref{fig:arch}c, a linear head shared by all sites maps each LoRA query state $\mathbf{z}_{s,g}$ to $c$ rank-1 pairs $(\hat{\mathbf{a}}_{s,g,i}, \hat{\mathbf{b}}_{s,g,i}) \in \mathbb{R}^{d} \times \mathbb{R}^{d}$, $i = 1, \dots, c$.
Module types differ in shape, so a per-type linear map lifts each normalized vector to the shape of its site:
\begin{align}
  \mathbf{a}_{s,g,i} &= U_{\tau} \operatorname{RMSNorm}(\hat{\mathbf{a}}_{s,g,i}) \in \mathbb{R}^{d^{\tau}_{\mathrm{in}}}, \\
  \mathbf{b}_{s,g,i} &= V_{\tau} \operatorname{RMSNorm}(\hat{\mathbf{b}}_{s,g,i}) \in \mathbb{R}^{d^{\tau}_{\mathrm{out}}},
\end{align}
where $U_{\tau} \in \mathbb{R}^{d^{\tau}_{\mathrm{in}} \times d}$ and $V_{\tau} \in \mathbb{R}^{d^{\tau}_{\mathrm{out}} \times d}$ are shared by all sites of type $\tau$.
The RMSNorm fixes the scale of each generated vector, so the decoder chooses directions and the per-type maps set magnitudes.
Stacking the $r = mc$ components of site $s$ as the rows of $A_s \in \mathbb{R}^{r \times d^{\tau}_{\mathrm{in}}}$ and the columns of $B_s \in \mathbb{R}^{d^{\tau}_{\mathrm{out}} \times r}$ gives
\begin{equation}
  \Delta W_s = B_s A_s = \sum_{g=1}^{m} \sum_{i=1}^{c} \mathbf{b}_{s,g,i} \mathbf{a}_{s,g,i}^{\top},
\end{equation}
with the constant LoRA scale $\alpha/r$ absorbed into $B_s$.
We initialize $V_{\tau}$ to zero, so $\Delta W = 0$ at the start and the interpreter begins at its pretrained behavior.

\paragraph{Optional prefix head.}
The prefix queries open a second channel into the interpreter, also depicted in \Cref{fig:arch}c.
A small MLP maps each final prefix state to an interpreter input embedding,
\begin{equation}
  \mathbf{p}_{p} = \mathbf{b}_{0} + W_{2} \operatorname{GELU}\big(W_{1} \operatorname{LN}(\mathbf{z}^{\mathrm{pre}}_{p})\big) \in \mathbb{R}^{d},
\end{equation}
where $\mathbf{b}_{0}$ is a bias shared by all positions, initialized to the embedding of a placeholder token, and $\mathbf{P}$ stacks $\mathbf{p}_{1}, \dots, \mathbf{p}_{P}$ as its rows.
The prefix queries are decoded jointly with the LoRA queries, so the two channels can split the content between them.
The same model can also run with the prefix left out, operating in a weights-only mode.

\paragraph{Interpreter.}
The interpreter in \Cref{fig:arch}d is a frozen VLM.
We install $\Delta W$ in every attention projection (including those of linear-attention layers in hybrid models) and every MLP projection of all layers, so each adapted linear layer computes
\begin{equation}
  (W_s + \Delta W_s)\, \mathbf{u} = W_s \mathbf{u} + B_s (A_s \mathbf{u})
\end{equation}
on its input $\mathbf{u}$.
This is ordinary LoRA, except that $A_s$ and $B_s$ are generated for each video instead of trained.
The interpreter's input is the prefix followed by the text query, $(\mathbf{P}, x)$, or the query $x$ alone when the prefix is left out.
It generates $y$ autoregressively as in \cref{eq:overview}.
$\Delta W_s$ is never materialized, and a compiled video is stored as $\sum_{s \in \mathcal{S}} r\,(d^{\tau}_{\mathrm{in}} + d^{\tau}_{\mathrm{out}})$ LoRA entries plus $P \cdot d$ prefix entries.

\paragraph{Trainable parameters.}
Only the compiler's LoRA, the readout embeddings, and the weight decoder, including both heads, are trained.
The vision encoder, the compiler's base weights, and the whole interpreter stay frozen.

\subsection{Training Objective}
\label{sec:objective}
If the prefix is always present during training, the generated weights learn to work only together with the tokens, and removing the tokens at test time hurts accuracy sharply.
We therefore train each example in two views that share one compilation.
The compiler and the decoder run once and produce $(\Delta W, \mathbf{P})$.
The interpreter then reads the conversation twice with the same $\Delta W$, once as $x$ alone and once as $(\mathbf{P}, x)$.
The loss combines the two views:
\begin{equation}
  \begin{split}
    \mathcal{L} = {} & (1-\lambda)\,\mathcal{L}_{\mathrm{CE}}\big(y \mid W + \Delta W,\, x\big) \\
                     & + \lambda\,\mathcal{L}_{\mathrm{CE}}\big(y \mid W + \Delta W,\, \mathbf{P}, x\big),
  \end{split}
  \label{eq:dual}
\end{equation}
where $\mathcal{L}_{\mathrm{CE}}$ is the cross-entropy on the target tokens and $\lambda = 0.5$.
Both terms train the decoder and the compiler through $\Delta W$, and the prefix tokens receive gradient only from the second term.
The first term makes the weights carry the video on their own.
The second term teaches the tokens to add what the weights miss.
Only the small interpreter runs twice, so the extra training cost is modest.
We use \cref{eq:dual} in both training stages described in \Cref{sec:data}.

\begin{table}[t]
  \caption{Evaluation benchmarks. All are multiple choice. Dur.\ is the mean video length in seconds.}
  \label{tab:benchmarks}
  \centering
  \small
  \setlength{\tabcolsep}{4pt}
  \begin{tabular}{llrrr}
    \toprule
    Benchmark & Split & \#Videos & \#Q & Dur. \\
    \midrule
    EgoSchema      & 500 subset      & 500   & 500   & 180   \\
    NExT-QA        & test            & 1,000 & 8,564 & 40    \\
    LongVideoBench & val             & 753   & 1,337 & 732   \\
    Video-MME      & full            & 900   & 2,700 & 1,021 \\
    MLVU           & dev, 7 MC tasks & 1,122 & 2,174 & 651   \\
    MVBench        & 20 tasks        & 3,871 & 4,000 & 14    \\
    \bottomrule
  \end{tabular}
\end{table}

\begin{table*}[t]
  \caption{Main results (accuracy, \%). Methods are grouped by how the video reaches the interpreter. \#Tokens is the number of tokens read in place of the video (full: up to 8,192 visual tokens). Among non-references, \textbf{bold} is the best and \underline{underline} the second best.}
  \label{tab:main_results}
  \centering
  \small
  \setlength{\tabcolsep}{5pt}
  \begin{tabular}{lcccccccc}
    \toprule
    Method & \#Tokens & EgoSchema & NExT-QA & LongVideoBench & Video-MME & MLVU & MVBench & Avg. \\
    \midrule
    \multicolumn{9}{l}{\textit{References}} \\
    SFT (0.8B)             & full & 66.0 & 75.3 & 52.4 & 57.5 & 63.2 & 58.8 & 62.2 \\
    In-context (0.8B)      & full & 48.6 & 69.4 & 52.2 & 53.9 & 61.5 & 54.7 & 56.7 \\
    In-context (9B)        & full & 72.0 & 84.8 & 64.8 & 69.4 & 76.7 & 71.4 & 73.2 \\
    Blind (0.8B)           & 0    & 22.0 & 32.0 & 32.1 & 32.7 & 35.8 & 31.5 & 31.0 \\
    \midrule
    \multicolumn{9}{l}{\textit{Token only}} \\
    KiToke (0.8B)          & 64 & 29.2 & 57.3 & 43.8 & 38.2 & 47.1 & 45.3 & 43.5 \\
    UniComp (0.8B)         & 64 & 24.4 & 51.3 & 39.4 & 34.7 & 42.5 & 44.2 & 39.4 \\
    Captioner (9B $\rightarrow$ 0.8B) & 1024 & 45.4 & 54.0 & 35.3 & 45.3 & 44.9 & 44.3 & 44.9 \\
    Compressor (9B $\rightarrow$ 0.8B) & 64 & 53.0 & 71.4 & \underline{46.8} & \underline{52.9} & 59.6 & \underline{58.7} & 57.1 \\
    \midrule
    \multicolumn{9}{l}{\textit{Weights only}} \\
    Doc2LoRA (0.8B $\rightarrow$ 0.8B) & 0  & 52.8 & 64.4 & 41.4 & 42.0 & 53.7 & 48.8 & 50.5 \\
    \rowcolor{gray!12} VaW (0.8B $\rightarrow$ 0.8B) & 0  & 58.8 & 67.5 & 42.2 & 44.3 & 55.0 & 52.3 & 53.4 \\
    \rowcolor{gray!12} VaW (9B $\rightarrow$ 0.8B)   & 0  & \underline{63.4} & 68.8 & 41.5 & 48.2 & 56.7 & 56.0 & 55.8 \\
    \midrule
    \multicolumn{9}{l}{\textit{Parametric and token memory}} \\
    \rowcolor{gray!12} VaW (0.8B $\rightarrow$ 0.8B) & 64 & 61.8 & \underline{71.5} & \textbf{47.1} & 52.2 & \underline{60.2} & 57.5 & \underline{58.4} \\
    \rowcolor{gray!12} VaW (9B $\rightarrow$ 0.8B)   & 64 & \textbf{66.4} & \textbf{75.7} & 45.8 & \textbf{54.9} & \textbf{61.2} & \textbf{62.1} & \textbf{61.0} \\
    \bottomrule
  \end{tabular}
\end{table*}

\section{Data and Evaluation}
\label{sec:data}

\subsection{Training Data}
We train VaW in two stages: pre-training on long-form video descriptions and text, then fine-tuning on visual instruction data.
Every example pairs a \emph{context} (a video, an image, or a text passage) with a conversation about it.
The compiler reads only the context and the interpreter reads only the conversation, so the context reaches the interpreter only through $\Delta W$ and $\mathbf{P}$.

\paragraph{Pre-training.}
The pre-training corpus has 1.0M examples, half video and half text.
For video, Qwen3.5-9B writes long-form descriptions of FineVideo \cite{farre2024finevideo} videos, such as dense descriptions, timestamped chronologies, and summaries, in proportions that follow the task types of Video-MME \cite{fu2025video} without using its data.
For text, each FineWeb-Edu \cite{penedo2024fineweb} document gives a \emph{reconstruction} example, where the interpreter reproduces the passage the compiler read, and a \emph{continuation} example, where it writes the text that follows.

\paragraph{Fine-tuning.}
The fine-tuning mix has 4.8M conversations, 63\% on videos and 37\% on images.
Conversations come from Cambrian-S-3M \cite{yang2025cambrians} for video and from a subset of Cambrian-7M \cite{tong2024cambrian} for image.
Open-ended answers on videos are written by Qwen3.5-9B, and all other targets are the datasets' ground truth.
Questions about the same video are grouped into multi-turn conversations, so one compiled video serves several questions.
Details of the data creation pipeline, task proportions, and prompts are in \Cref{app:data}.

\subsection{Evaluation Benchmarks}
\label{sec:eval}
We evaluate on six multiple-choice video QA benchmarks whose videos range from seconds to hours (\Cref{tab:benchmarks}): EgoSchema \cite{Mangalam2023EgoSchema}, NExT-QA \cite{xiao2021next}, LongVideoBench \cite{wu2024longvideobench}, Video-MME \cite{fu2025video}, MLVU \cite{zhou2024mlvu}, and MVBench \cite{Li2023MVBenchAC}.

\paragraph{Protocol.}
All methods use 128 uniformly sampled frames with at most 128 visual tokens each, i.e., at most 8,192 visual tokens per video, and no subtitles or audio.
Each question is shown with its lettered options.
We decode greedily (up to 256 tokens), parse the predicted letter, and report accuracy over all questions.
For VaW, the compiler reads only the video, and the question and options go only to the interpreter.

\subsection{Baselines}
All baselines and references share the same Qwen3.5 backbone family as VaW, using Qwen3.5-0.8B as the small model and Qwen3.5-9B as the larger teacher or compiler.

\paragraph{References.}
\emph{In-context} evaluates the 0.8B and 9B models zero-shot on all visual tokens of the 128 frames.
\emph{SFT} fine-tunes the 0.8B interpreter with the full video in context on our fine-tuning mix using a static LoRA of rank 64, four times the rank of VaW's generated LoRA.
\emph{Blind} evaluates the 0.8B model on the question and options alone without any video input.

\paragraph{Token compression.}
\emph{KiToke} \cite{huang2026kitokekernelbasedintervalawaretoken} and \emph{UniComp} \cite{yuan2026unicomp} are training-free methods that prune the visual tokens of the in-context 0.8B interpreter to at most 64 tokens for the whole video.
\emph{Captioner} uses the 9B model to write a question-agnostic video description of up to 1{,}024 tokens, which the 0.8B model then reads in place of visual tokens.
\emph{Compressor} uses the same 9B compiler as VaW, but passes the video to the 0.8B interpreter solely as 64 prefix tokens with no generated weights.

\paragraph{Weight generation.}
\emph{Doc2LoRA} follows the layer-wise design of Doc-to-LoRA \cite{charakorn2026doctolora} and Frames2LoRA \cite{suri2026frames2loraparametricvideointernalization}.
It keeps the setup of VaW with a 0.8B compiler and $P = 0$ but generates the LoRA of interpreter layer $j$ only from compiler layer $j$, at rank 64.

\section{Results and Discussions}

\subsection{Main Results}
We pair the Qwen3.5-0.8B interpreter with a Qwen3.5-0.8B or Qwen3.5-9B compiler, each with $P = 0$ (weights only) and $P = 64$.
Details of VaW and the baselines are in \Cref{app:impl,app:baselines}.

\Cref{tab:main_results} compares VaW with the baselines.
Even with the 0.8B compiler, VaW (0.8B $\rightarrow$ 0.8B) with $P = 64$ outperforms the same interpreter reading the full video in context by 1.7 points (58.4 vs.\ 56.7) while reading 64 tokens instead of up to 8,192.
Pairing the interpreter with the 9B compiler pushes accuracy to 61.0, outperforming full in-context reading by 4.3 points and coming within 1.2 points of SFT, which is fine-tuned with the full video in context.
Training-free pruning does not hold up at this budget, as KiToke and UniComp fall 13 to 17 points below in-context reading.
A 9B$\rightarrow$0.8B question-agnostic caption of 1{,}024 tokens is similarly weak, at 44.9 average.

\paragraph{All-Layer vs. Layer-Wise Decoding}
Doc2LoRA shares the compiler, readout tokens, LoRA sites, data, and schedule of VaW (0.8B $\rightarrow$ 0.8B) with $P = 0$, and differs in its layer-wise decoder and its LoRA rank of 64, four times that of VaW.
This setting favors the layer-wise design, since the compiler and the interpreter have the same depth and their layers match one to one.
All-layer decoding still wins on all six benchmarks, by 2.9 points on average and by up to 6.0 points on EgoSchema.
\Cref{sec:weight_analysis} shows why: every interpreter layer relies mostly on the upper compiler layers rather than on its counterpart, so a layer-wise decoder cuts each layer off from the features it uses.
Without a fixed layer mapping, the compiler and the interpreter also no longer need the same depth.

\paragraph{Strong-to-Weak Transfer}
Because no layer correspondence is imposed, a 32-layer 9B compiler can write videos into the 24-layer 0.8B interpreter.
Compared with a 0.8B compiler, it improves the same interpreter by 2.6 points with $P = 64$ and by 2.4 points with $P = 0$, and with $P = 64$ the gain holds on every benchmark except LongVideoBench.
With only 64 tokens, the interpreter even surpasses its own fine-tuned, full-context accuracy on EgoSchema and MVBench.
The 9B compiler runs once per video and never sees the question, so every question is answered by the 0.8B interpreter alone.

\paragraph{Generated Weights Help}
With the same 9B compiler and 64 tokens, VaW improves over the 9B$\rightarrow$0.8B compressor, which passes the video only as tokens, by 3.9 points, with the largest gain on EgoSchema.
The compressor also differs in its readout and compiler LoRA (\Cref{app:baselines}), but a tokens-only model also falls 5.1 points below VaW trained with the same recipe (\Cref{tab:ablation}).
Even with no tokens at all, VaW (9B $\rightarrow$ 0.8B) outperforms both pruning methods with 64 tokens by more than 11 points, and it beats in-context reading of the full video on EgoSchema and MVBench.

\begin{table*}[t]
  \caption{\textbf{Ablations of VaW} (accuracy, \%). Each group compares against its gray row, and $\Delta$ is the change in average accuracy from it. Groups differ slightly in training recipe (\Cref{app:ablation}). $^{\dagger}$Trained with the recipe of the tokens-only model, so the gray row differs from \Cref{tab:main_results}. $^{\ddagger}$The design of Doc-to-LoRA, with 24 of the 32 compiler layers matched to the 24 interpreter layers.}
  \label{tab:ablation}
  \centering
  \small
  \setlength{\tabcolsep}{5pt}
  \begin{tabular}{lcccccccc}
    \toprule
    Setting & EgoSchema & NExT-QA & LongVideoBench & Video-MME & MLVU & MVBench & Avg. & $\Delta$ \\
    \midrule
    \multicolumn{9}{l}{\textit{Channels at test time (9B $\rightarrow$ 0.8B, $r = 16$)}} \\
    \rowcolor{gray!12} Both (\Cref{tab:main_results}) & 66.4 & 75.7 & 45.8 & 54.9 & 61.2 & 62.1 & 61.0 & -- \\
    Tokens left out & 58.0 & 69.0 & 39.9 & 46.2 & 55.0 & 55.9 & 54.0 & $-$7.0 \\
    LoRA left out & 40.2 & 59.2 & 39.1 & 46.0 & 50.2 & 48.7 & 47.2 & $-$13.8 \\
    Another video's LoRA & 60.8 & 72.0 & 44.9 & 52.4 & 58.5 & 59.4 & 58.0 & $-$3.0 \\
    Another video's tokens & 52.6 & 62.1 & 40.9 & 44.8 & 52.7 & 47.9 & 50.2 & $-$10.8 \\
    \midrule
    \multicolumn{9}{l}{\textit{Weights in training (9B $\rightarrow$ 0.8B, $r = 16$)$^{\dagger}$}} \\
    \rowcolor{gray!12} Weights and tokens & 67.2 & 75.7 & 45.6 & 53.8 & 61.7 & 61.2 & 60.9 & -- \\
    Tokens only & 49.8 & 70.3 & 46.6 & 51.9 & 59.9 & 56.2 & 55.8 & $-$5.1 \\
    \midrule
    \multicolumn{9}{l}{\textit{Decoder (9B $\rightarrow$ 0.8B, $P = 0$, $r = 64$)}} \\
    \rowcolor{gray!12} All-layer & 58.4 & 69.4 & 43.3 & 47.6 & 54.4 & 55.6 & 54.8 & -- \\
    Layer-wise, depth-matched$^{\ddagger}$ & 55.6 & 66.0 & 42.4 & 43.7 & 54.3 & 50.8 & 52.1 & $-$2.7 \\
    \midrule
    \multicolumn{9}{l}{\textit{Prefix tokens (9B $\rightarrow$ 0.8B, $r = 64$)}} \\
    $P = 16$ & 64.6 & 73.2 & 46.1 & 50.3 & 58.2 & 58.6 & 58.5 & $-$2.5 \\
    \rowcolor{gray!12} $P = 64$ & 67.6 & 76.0 & 46.2 & 54.0 & 60.9 & 61.2 & 61.0 & -- \\
    $P = 128$ & 67.8 & 76.1 & 47.9 & 55.2 & 62.4 & 61.9 & 61.9 & $+$0.9 \\
    \bottomrule
  \end{tabular}
\end{table*}

\subsection{Ablation Study}
\label{sec:ablation}
\begin{figure*}[t]
  \centering
  \includegraphics[width=\textwidth]{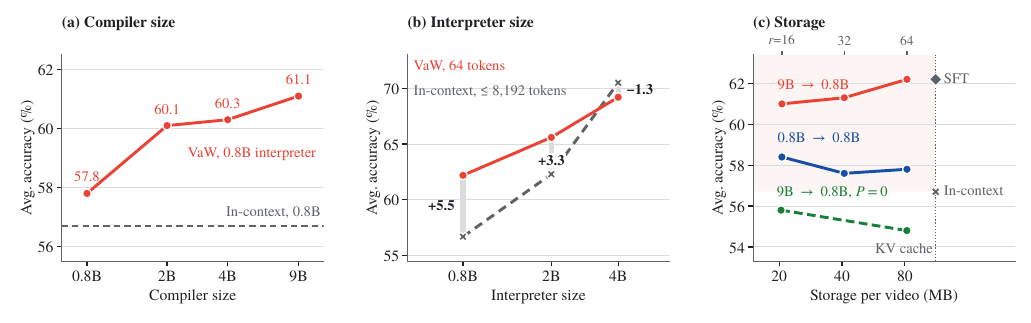}
  \caption{\textbf{Scaling VaW} (average accuracy over the six benchmarks). \textbf{(a)} Compilers of different sizes writing for the same 0.8B interpreter, against that interpreter reading the full video in context. \textbf{(b)} Interpreters of different sizes, with VaW reading 64 tokens and in-context reading up to 8,192. Labels give the gap. \textbf{(c)} Storage of a compiled video as the LoRA rank $r$ varies (top axis), with $P = 64$ prefix tokens unless noted. Gray marks read the video in context and keep its KV cache (112.4 MB, \Cref{app:compute}), and the shaded region beats in-context reading with less memory.}
  \label{fig:scaling}
\end{figure*}

\Cref{tab:ablation} ablates the design of VaW on all six benchmarks, and \Cref{fig:scaling} varies the size of the compiler, the interpreter, and the compiled video.

\paragraph{What Each Channel Carries}
VaW needs both channels at inference: leaving out the tokens costs 7.0 points, and leaving out the LoRA costs 13.8.
Both channels are specific to the video.
Answering with another video's LoRA, while keeping this video's tokens, lowers accuracy on every benchmark, by 3.0 points on average, and another video's tokens cost 10.8.
The tokens thus carry most of the video, and the weights add to it.
The weights are also worth training, as a model trained with tokens alone falls 5.1 points below VaW trained with the same recipe, and 17.4 points on EgoSchema.
On Video-MME, the tokens help most on fine-grained appearance and on short videos, while the weights alone already answer summary questions (\Cref{app:video_mme}).

\paragraph{Decoder}
A layer-wise decoder, the design of Doc-to-LoRA, generates the LoRA of each interpreter layer from one compiler layer.
With the 9B compiler, we match 24 of its 32 layers to the 24 interpreter layers.
This still falls 2.7 points below all-layer decoding and loses on every benchmark, in line with the 0.8B comparison in \Cref{tab:main_results}, so the gain comes from removing the fixed mapping rather than from a mismatch in depth.

\paragraph{Prefix Tokens}
Accuracy grows with the token budget on every benchmark, as 16 tokens lose 2.5 points against 64, and 128 tokens add another 0.9.
Each token takes only 2 KB of storage but lengthens the interpreter's input for every question, so we keep $P = 64$.

\paragraph{Compiler and Interpreter Size}
With the same 0.8B interpreter, accuracy grows steadily with the compiler, from 57.8 with 0.8B to 61.1 with 9B (\Cref{fig:scaling}a).
Larger interpreters also work with VaW (\Cref{fig:scaling}b).
It beats in-context reading of the full video by 5.5 points at 0.8B and 3.3 points at 2B, and comes within 1.3 points at 4B, while reading 64 tokens instead of up to 8,192.
The gain is largest for a weak interpreter, where a strong compiler has the most to add.

\paragraph{Storage}
The size of a compiled video is set by the LoRA rank and the number of prefix tokens, not by the length of the video.
Accuracy barely depends on the rank (\Cref{fig:scaling}c).
From $r = 16$ to 64, storage grows fourfold, while accuracy changes by $+$1.2 points with the 9B compiler, $-$0.6 with the 0.8B compiler, and $-$1.0 without tokens.
We therefore use $r = 16$, which stores a compiled video in 20.6 MB, 5.5 times less than the interpreter's 112.4 MB KV cache of the video (\Cref{app:compute}), while answering more accurately than reading the video in context.
A compiled video is a reusable per-video state much like this cache, but whereas the KV cache grows linearly with the number of frames, a compiled video stays fixed in size.

\subsection{Analysis of Generated Weights}
\label{sec:weight_analysis}

We analyze VaW (9B $\rightarrow$ 0.8B) with $P = 64$ and with $P = 0$ from \Cref{tab:main_results} on the 900 Video-MME videos.

\begin{figure}[htb]
  \centering
  \includegraphics[width=\columnwidth,trim={0 9 0 0},clip]{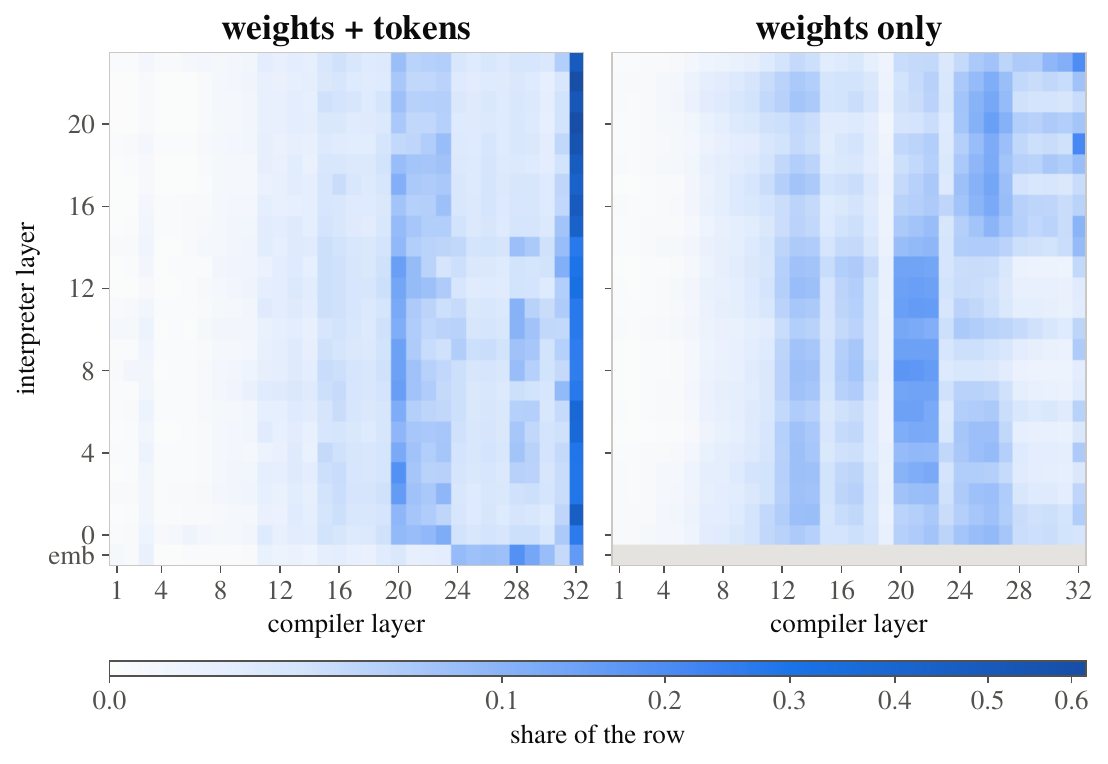}
  \caption{\textbf{Compiler layers read by the weights}, for VaW (9B $\rightarrow$ 0.8B) with $P = 64$ (weights + tokens) and $P = 0$ (weights only) on Video-MME. Each cell shows the share of an interpreter layer's video-specific update attributable to a compiler layer (\Cref{app:weight_analysis}), and each row sums to one. The ``emb'' row is the prefix tokens.}
  \label{fig:layer_corr}
\end{figure}

\paragraph{Where the Weights Read From}
VaW does not tie interpreter layers to compiler layers, so we measure how strongly each interpreter layer depends on each compiler layer by replacing a compiler layer's readout states with their cross-video average and recording the resulting change in the generated LoRA (\Cref{fig:layer_corr}; formal metric in \Cref{app:weight_analysis}).

Two patterns stand out.
First, the decoder relies mostly on the upper half of the compiler: compiler layers 17 to 32 account for 88\% to 94\% of every row with prefix tokens and 65\% to 86\% without them.
Second, the pattern depends on the prefix tokens.
With them, every interpreter layer relies mainly on the same two compiler layers, 32 and 20, so all layers receive one shared summary of the video while the prefix tokens draw from layers 24 to 32.
Without them, deeper interpreter layers rely on progressively deeper compiler layers, which forms a loose diagonal.
Neither model learns a one-to-one mapping between layers, which is why a compiler and an interpreter of different depths can be paired.

\begin{figure}[t]
  \centering
  \includegraphics[width=\columnwidth,trim={0 9 0 0},clip]{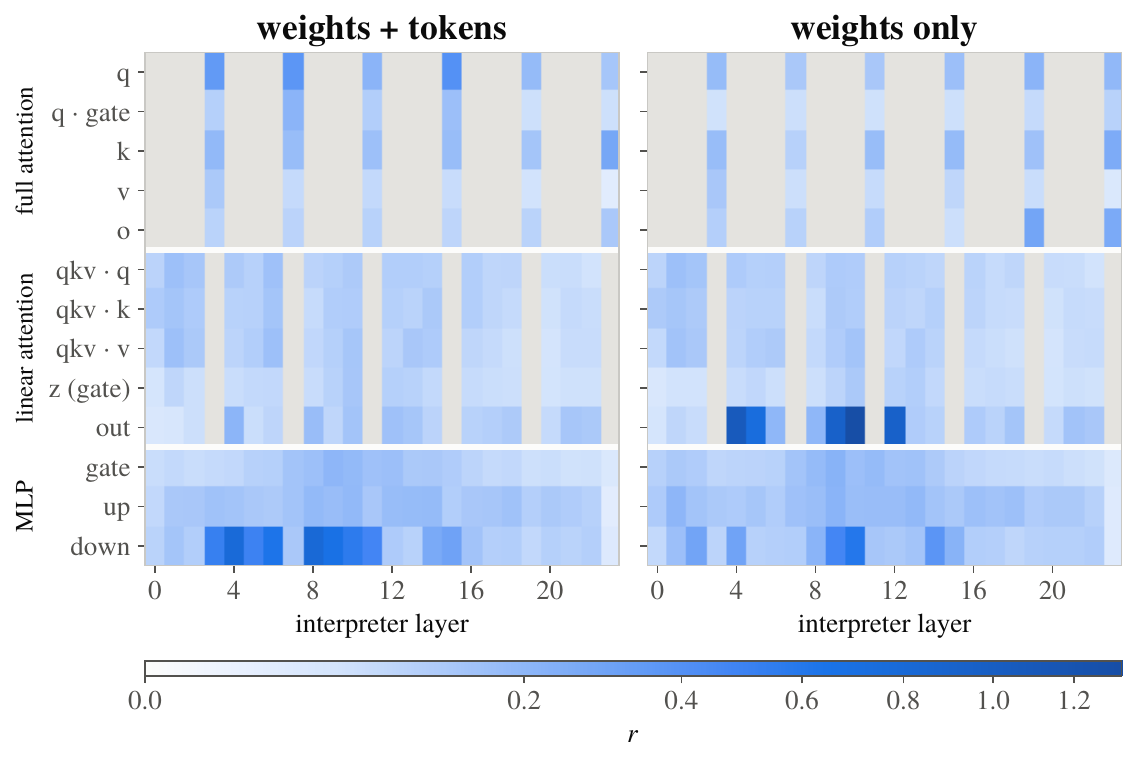}
  \caption{\textbf{Interpreter modules changed by the weights}, for VaW (9B $\rightarrow$ 0.8B) with $P = 64$ (weights + tokens) and $P = 0$ (weights only) on Video-MME. Each cell shows the relative output change $r_s$ induced by the generated LoRA (\Cref{app:weight_analysis}). Gray cells mark modules that the layer does not have.}
  \label{fig:module_change}
\end{figure}

\paragraph{Where the Weights Write To}
We also measure which interpreter modules the generated LoRA perturbs most (\Cref{fig:module_change}), comparing the output norm of each module's LoRA delta against that of its frozen weight across question tokens (\Cref{app:weight_analysis}).
With prefix tokens, the weights mostly change the full-attention queries and the MLP down projections, telling the interpreter where to look in the prefix while keeping a baseline summary in the weights.
Without prefix tokens, the largest changes shift more than fourfold to the linear-attention output projections and double on the full-attention output projections, which write directly into the residual stream to carry the video content themselves.
This division of labor matches the task breakdown on Video-MME (\Cref{app:video_mme}): the weights alone already handle global synopsis questions, whereas adding the prefix tokens boosts OCR by more than 20 points and object recognition by nearly 14.

\section{Conclusion and Future Work}

We introduced Video-as-Weights, which compiles a video once into the parametric and token memory (LoRA weights and optional prefix tokens) of a frozen interpreter VLM.
A single decoder writes both channels and reads every compiler layer, supporting both same-model compilation and strong-to-weak transfer from a larger compiler to a smaller interpreter.
Storing each video in 20.6 MB ($5.5\times$ less than the interpreter's KV cache) with 64 tokens, VaW outperforms full in-context reading with both a 0.8B and a 9B compiler.
Our analysis shows that the weights carry the gist of a video and the tokens add its details.

Three directions follow from these results.
First, the gain from tokens shrinks on long videos, which suggests scaling the token budget and the LoRA rank with video length.
Second, compiled videos could be composed, merging the LoRAs of several clips or videos into a single memory that the interpreter can query.
Third, the same compiler--interpreter design could extend to larger interpreters and to other modalities such as audio.

\newpage
\section*{Acknowledgments}

This work benefited greatly from the collective insights, dedication, and collaboration of all authors, whose thoughtful feedback helped shape every stage of the project.
We gratefully acknowledge the University of Waterloo for financial support, and Google for providing both the funding and the computational resources essential to this research.

{
    \small
    \bibliographystyle{ieeenat_fullname}
    \bibliography{main}
}

\clearpage
\onecolumn
{\centering
  \Large
  \textbf{\thetitle}\\
  \vspace{0.5em}Appendix \\
  \vspace{1.0em}
  \par}
\appendix

\section{Implementation Details}
\label{app:impl}

We use Qwen3.5-0.8B as the interpreter and Qwen3.5-0.8B or Qwen3.5-9B as the compiler.
\Cref{tab:hparams} lists the hyperparameters of VaW.
The generated LoRA adapts all 150 linear projections in the interpreter's attention, linear-attention, and MLP blocks, and the compiler's LoRA covers the same module types.
Each stage trains for one epoch with cross-entropy on the target tokens, using micro-batches packed up to a fixed token budget, a linear warmup, and cosine decay.
For the 9B compiler, we keep the readout embeddings at their initialization.
Models with a 0.8B compiler train on one node of eight H100 GPUs, and models with a 9B compiler train on two nodes.
Data creation details are in \Cref{app:data}, and baseline details are in \Cref{app:baselines}.

\begin{table}[h]
  \caption{VaW hyperparameters. Pairs separated by ``/'' are for the 0.8B / 9B compiler.}
  \label{tab:hparams}
  \centering
  \small
  \setlength{\tabcolsep}{4pt}
  \begin{tabular}{lcc}
    \toprule
    \textbf{Architecture}                    &  &  \\
    \midrule
    Compiler layers $L$                      & \multicolumn{2}{c}{24 / 32} \\
    Interpreter hidden size $d$              & \multicolumn{2}{c}{1024} \\
    Readout tokens $K$                       & \multicolumn{2}{c}{32 to 512} \\
    Readout ratio $\rho$                     & \multicolumn{2}{c}{8} \\
    Decoder $d_h$, $N$, heads                & \multicolumn{2}{c}{1024, 6, 8} \\
    LoRA rank $r$, scale $\alpha$            & \multicolumn{2}{c}{16, 32} \\
    Components per query $c$                 & \multicolumn{2}{c}{4} \\
    Adapted sites $|\mathcal{S}|$            & \multicolumn{2}{c}{150} \\
    Compiler LoRA rank $r_c$, scale          & \multicolumn{2}{c}{32, 64} \\
    Prefix tokens $P$                        & \multicolumn{2}{c}{0 or 64} \\
    \midrule
    \textbf{Training}                        & Pre-training & Fine-tuning \\
    \midrule
    Examples                                 & 1.0M & 4.8M \\
    Steps (one epoch)                        & 9,252 & 14,287 \\
    Micro-batches per step                   & 32 & 32 \\
    Tokens / rows per micro-batch            & 12,288 / 12 & 32,768 / 16 \\
    Frames per video                         & 64 & $\le$128, $\le$1 fps \\
    Peak learning rate                       & $1 \times 10^{-4}$ & $1 \times 10^{-4}$$^{*}$ \\
    Warmup steps                             & 1,200 & 450 \\
    Final / peak learning rate               & 0.2 / 0.1 & 0.01 \\
    Optimizer                                & \multicolumn{2}{c}{AdamW, $\beta = (0.9, 0.95)$} \\
    Weight decay, grad.\ clipping            & \multicolumn{2}{c}{0.01, 1.0} \\
    Precision                                & \multicolumn{2}{c}{bf16} \\
    \bottomrule
  \end{tabular}
\end{table}

\section{Data Creation Details}
\label{app:data}

\Cref{tab:data_tasks} lists the task families, sampling proportions, and example prompts for pre-training and fine-tuning.
In both stages, the compiler reads only the visual or text context with the readout tokens, and the interpreter reads only the conversation.

\begin{table}[t]
  \caption{Task families, sampling proportions, and example prompts for pre-training and fine-tuning. ``Modality (\%)'' is the sampling share within video or text during pre-training, and ``Total (\%)'' is the share of the 1.0M pre-training corpus.}
  \label{tab:data_tasks}
  \centering
  \small
  \setlength{\tabcolsep}{6pt}
  \begin{tabular}{lccp{8.8cm}}
    \toprule
    Task family & Modality (\%) & Total (\%) & Example prompt \\
    \midrule
    \multicolumn{4}{l}{\textit{Pre-training: Video descriptions (FineVideo, 50.0\% of corpus)}} \\
    People and objects & 16.1 &  8.1 & \emph{List the people and objects that appear in this video, and when each is on screen.} \\
    Dense description  & 14.3 &  7.2 & \emph{Describe this video in detail, covering what is shown from beginning to end. Do not use timestamps.} \\
    Dense chronology   & 14.1 &  7.1 & \emph{Walk through the video in order, one line per moment you can actually see. Start each line with its time in seconds in square brackets, like [142.5s].} \\
    Summary            & 13.4 &  6.7 & \emph{Summarise this video: what it is, what it covers, and how it is put together.} \\
    Action detail      & 11.1 &  5.6 & \emph{Describe the actions in this video in detail: what is done, and how.} \\
    Change over time   &  8.8 &  4.4 & \emph{Describe how this video changes as it goes on: the setting, the mood, the subject matter.} \\
    Event order        &  5.6 &  2.8 & \emph{List the events of this video in the order they happen. No timestamps: the order itself is the point.} \\
    Scene localization &  5.6 &  2.8 & \emph{State when each notable thing happens in this video, giving the time in seconds for each one.} \\
    Setting and style  &  5.5 &  2.8 & \emph{Describe the setting of this video and the way it is shot and cut.} \\
    On-screen text     &  5.4 &  2.7 & \emph{Report the text that appears on screen in this video, and where it appears.} \\
    \midrule
    \multicolumn{4}{l}{\textit{Pre-training: Text tasks (FineWeb-Edu, 50.0\% of corpus)}} \\
    Reconstruction     & 50.0 & 25.0 & \emph{Reproduce the text exactly as it was written.} \\
    Continuation       & 50.0 & 25.0 & \emph{Keep writing from where the text stopped.} \\
    \midrule
    \multicolumn{4}{l}{\textit{Fine-tuning: Visual instruction tuning (Cambrian-S-3M and Cambrian-7M)}} \\
    Open-ended QA      & --   & --   & \emph{Use the video to answer the question below. [Question] Say what in the video shows this, then give your answer.} \\
    Short QA           & --   & --   & \emph{Watch the video and answer the question below. [Question] Answer in one or two sentences.} \\
    Multiple choice    & --   & --   & \emph{Answer this multiple-choice question about the video. [Question] [Options]} \\
    Captioning         & --   & --   & \emph{[Question] Write about 150--250 words. Write plain prose.} \\
    \bottomrule
  \end{tabular}
\end{table}

\subsection{Pre-training Data}

The pre-training corpus has 1.0M single-turn examples, made up of 504K video examples from 43.8K FineVideo \cite{farre2024finevideo} videos and 500K text examples from 250K FineWeb-Edu \cite{penedo2024fineweb} documents.

\paragraph{Video task proportions.}
We group video descriptions into the 10 task families in \Cref{tab:data_tasks}.
To choose their sampling proportions without using benchmark data, we relate the 10 families to the 12 question categories of Video-MME \cite{fu2025video} through a hand-specified coverage matrix $C \in \mathbb{R}^{12 \times 10}$.
Let $\mathbf{w}$ be the category distribution of Video-MME: 16.8\% object reasoning, 13.1\% object recognition, 12.0\% information synopsis, 11.6\% action recognition, 10.6\% action reasoning, 9.9\% counting, 8.2\% attribute perception, 6.6\% temporal reasoning, 5.1\% OCR, 2.1\% spatial reasoning, 2.0\% temporal perception, and 2.0\% spatial perception.
We find the family probabilities $\mathbf{p} \in \Delta^{9}$ by solving
\begin{equation}
  \min_{\mathbf{p} \in \Delta^{9},\; \mathbf{p} \ge \mathbf{f}_{\min}}
  \frac{1}{2}\|C\mathbf{p} - \mathbf{w}\|_2^2
  + \frac{\lambda_{\mathrm{ridge}}}{2}\|\mathbf{p} - \mathbf{u}\|_2^2,
  \label{eq:task_shares}
\end{equation}
where $\mathbf{u} = \frac{1}{10}\mathbf{1}$ is the uniform distribution, $\lambda_{\mathrm{ridge}} = 0.10$, and $\mathbf{f}_{\min}$ sets a minimum share of 15\% for dense chronology and 6\% each for event order and scene localization before final normalization.
The resulting shares are listed in \Cref{tab:data_tasks}.

\paragraph{Video prompts and teacher generation.}
Videos whose short side exceeds 384 pixels are resized to 384 pixels while keeping their aspect ratio and frame timestamps, and each video is sampled at 64 frames with at most 128 visual tokens per frame.
We draw up to 12 tasks per video without repeating the same family and prompt wording.
To match the task to the video length, we divide videos into short clips under 120 seconds, medium clips of 120 to 480 seconds, and long clips of at least 480 seconds.
On long clips, we multiply the sampling weights of the five temporal families---dense chronology, summary, event order, change over time, and scene localization---by $1.25$ before re-normalizing, and on short clips we divide them by $1.25$.
On-screen text is sampled only when the video has a short side of at least 240 pixels and its scene metadata mentions visible text such as signs, captions, subtitles, title cards, logos, or screens.

Each prompt has four parts: one of six phrasings for the chosen family, a timestamp instruction for the two timed families, a target word-count range, and a style instruction.
For dense chronology and scene localization, the prompt asks the model to start every line with a bracketed time in seconds such as \texttt{[142.5s]} and to write at most 32 entries spaced at least $\max(\text{duration}/64, 1)$ seconds apart.
The target length is drawn from 150--250, 250--450, or 450--800 words, with longer ranges favored for dense chronology and scene localization, shorter ranges favored for summary and on-screen text, and uniform weights for the remaining families.
The style instruction asks for plain prose, flowing narration, or a one-item-per-line list, with timed families always using a list.
Qwen3.5-9B generates each description greedily with thinking disabled and a limit of 4,096 tokens.
If an output fails quality checks, we regenerate it once by sampling at temperature $0.7$, top-$p$ $0.8$, top-$k$ $20$, and presence penalty $1.0$.
We discard outputs that are shorter than 40 words, repeat 12-grams more than 2\% of the time, repeat lines more than 15\% of the time, fall below 95\% ASCII characters, mention metadata or transcripts, or take a question-and-answer format.
For the timed families, we also check that all timestamps lie within the video duration and increase monotonically for chronologies.

\paragraph{Text tasks.}
For text pre-training, we take FineWeb-Edu \cite{penedo2024fineweb} documents of at least 512 tokens, tokenize the first 40,000 characters with the Qwen3.5 tokenizer, and allow cuts only at sentence or paragraph boundaries.
Every kept document gives one reconstruction example and one continuation example, each accounting for 25.0\% of the pre-training mix.
In both tasks, the compiler reads a passage chosen from three length bands: 256--1,024 tokens, 1,024--2,048 tokens, or 2,048--3,584 tokens.
For reconstruction, the interpreter receives one of 16 prompts asking it to repeat the passage word for word.
For continuation, the interpreter receives one of 16 prompts asking it to continue the text, and the target is the next 256--3,584 tokens of the document.
If either span cannot be cut cleanly at sentence boundaries, we drop the document so the two tasks stay balanced.

\subsection{Fine-tuning Data}

The fine-tuning mix has 4,835,804 conversations across 2,175,625 videos and images, with 3.03M multi-turn conversations on videos and 1.81M single-turn conversations on images.

\paragraph{Sources and preprocessing.}
Video conversations come from Cambrian-S-3M \cite{yang2025cambrians}.
For its LLaVA-Video-178K \cite{zhang2024llavavideo} portion, we take the 186.8K videos and annotations directly from the original release, yielding 1.60M conversations and 5.20M turns, and remove those six source folders from Cambrian-S-3M to avoid duplicates.
The remaining Cambrian-S-3M sources provide 1.43M video conversations.
Many Cambrian-S-3M questions refer to a short segment inside a long recording; when start and end times are given, we cut out that segment as a clip of at least 1.0 second before sampling frames, so that the 128 sampled frames cover the queried event rather than the rest of the recording.
During training, videos are sampled at up to 128 frames and at most 1 fps, with at most 128 visual tokens per frame.
For images, we draw a source-stratified sample of up to 1.0M images and 1.81M questions from Cambrian-7M \cite{tong2024cambrian}, dropping text-only entries and images with aspect ratio above 200:1.
Each image is read as a single frame with up to 2,048 visual tokens.

\paragraph{Prompts and targets.}
We strip dataset-specific formatting instructions from all questions and sort turns into four groups: open-ended QA, short QA with ground-truth answers of at most five words, single- or multi-answer multiple choice with 2 to 10 options, and captioning requests with long descriptive answers.
On LLaVA-Video-178K, 81.0\% of turns are open-ended QA, 14.8\% are multiple choice, 3.4\% are captioning, and 0.8\% are short QA.
Each turn is formatted by pairing one of 8 to 10 task lead-ins with the question, any answer options, and a closing instruction:
\begin{itemize}
  \setlength{\itemsep}{1pt}
  \setlength{\topsep}{2pt}
  \item \textbf{Video open-ended and short QA.} Qwen3.5-9B writes the answers from 128 frames, using greedy decoding up to 2,048 tokens and one sampled retry if needed. On 70\% of turns, the prompt asks the model to cite the visual evidence before answering, and on the remaining 30\% it asks for a direct answer in one or two sentences. We apply the same quality filters as in pre-training and also reject replies that only echo frame timestamps.
  \item \textbf{Multiple choice, captioning, and image QA.} All other targets use the datasets' ground truth. For multiple-choice questions, we shuffle the options on every turn and format them in one of five ways: \texttt{A.\ text}, a list under an \texttt{Options:} header, \texttt{(A) text}, \texttt{A) text}, or a single inline list separated by semicolons. The target is the ground-truth letter and option text in the matching style. Captioning turns use the dataset's reference caption with a matching word-count prompt, falling back to the 9B teacher only if the reference fails the quality checks. Image turns from Cambrian-7M use the dataset's ground-truth answers directly.
\end{itemize}

\paragraph{Multi-turn conversations.}
Captioning turns stay as single-turn examples, while all other accepted turns for the same video are grouped into conversations of 2 to 6 turns, with sampling weights of $1.0, 1.6, 1.6, 1.0, 0.6$ for lengths 2 through 6 and an average of 3.3 turns per conversation on LLaVA-Video-178K.
Within each conversation, multiple-choice and short QA turns come before open-ended QA turns.
We stop adding turns once the earlier answers in the conversation reach 220 words, and we skip any turn whose answer shares more than 25\% of its 12-grams with an earlier answer in the same conversation.
Turns after the first open with a short transition phrase such as \emph{``Another question about the same video.''} 35\% of the time.

\section{Baseline and Reference Details}
\label{app:baselines}

All baselines and references follow the evaluation protocol of \Cref{sec:eval}: 128 uniformly sampled frames with at most 128 visual tokens each, greedy decoding of up to 256 tokens, and accuracy from the parsed option letter.
\Cref{tab:baseline_hparams} lists the training hyperparameters of the three trained baselines.

\begin{table}[h]
  \caption{Training hyperparameters of the trained baselines. All use AdamW with $\beta = (0.9, 0.95)$, weight decay 0.01, bf16, a linear warmup with cosine decay, and 32 micro-batches per step.}
  \label{tab:baseline_hparams}
  \centering
  \small
  \begin{tabular}{lcccc}
    \toprule
                                        & SFT (0.8B) & Compressor (9B$\rightarrow$0.8B) & \multicolumn{2}{c}{Doc2LoRA} \\
                                        &     &               & Pre-training & Fine-tuning \\
    \midrule
    Trained parameters                  & interpreter LoRA & compiler LoRA, MLP & \multicolumn{2}{c}{compiler LoRA, decoder} \\
    Initialization                      & Qwen3.5-0.8B & earlier pre-training & Qwen3.5-0.8B & its pre-training \\
    Data                                & fine-tuning mix & fine-tuning mix & video--text mix & fine-tuning mix \\
    Steps                               & 14,287 & 14,000 & 9,252 & 14,287 \\
    Peak learning rate                  & $1.1 \times 10^{-4}$ & $1.1 \times 10^{-4}$ & $1 \times 10^{-4}$ & $1.1 \times 10^{-4}$ \\
    Warmup steps                        & 450 & 450 & 1,200 & 450 \\
    Final / peak learning rate          & 0.01 & 0.01 & 0.2 & 0.01 \\
    Gradient clipping                   & 1.0 & 1.0 & 2.0 & 1.0 \\
    Max tokens / rows per micro-batch   & 32,768 / 16 & 32,768 / 16 & 12,288 / 12 & 32,768 / 16 \\
    GPUs (H100)                         & 8 & 16 & 8 & 8 \\
    \bottomrule
  \end{tabular}
\end{table}

\paragraph{In-context.}
Qwen3.5-0.8B or Qwen3.5-9B reads all visual tokens of the 128 frames followed by the question, served with vLLM at temperature 0 with thinking disabled.

\paragraph{SFT.}
The same model is fine-tuned on our fine-tuning mix with the video in context.
It trains a LoRA of rank 64 and scale 128 on the same 150 projections that VaW adapts, while the vision encoder and the base weights stay frozen.
It uses the same data loader and packing as VaW with a 0.8B compiler.

\paragraph{Blind.}
The base Qwen3.5-0.8B receives only the question and options, with no video and no LoRA.

\paragraph{KiToke.}
KiToke has no public code, so we implement it from the paper on top of Qwen3.5-0.8B.
It is training-free and does not see the question.
On the visual tokens of the whole video, it estimates a Gaussian-kernel density, keeps 64 tokens by pivotal sampling with inclusion probability inversely proportional to the density, and merges every other token into its nearest kept token within content-adaptive temporal intervals.
Kept tokens keep their original positions, and the timestamp and delimiter tokens of a frame group are kept only if the group keeps at least one visual token.
The paper's default kernel width and thresholds are $800, 110, 70,$ and $0.4$, which are off-scale for Qwen3.5-0.8B features where the kernel becomes nearly constant.
We therefore rescale them to the feature statistics of 100 unlabeled videos per benchmark and use the median across benchmarks: $\alpha = 17.5$, $\tau_{\mathrm{diff}} = 6.2$, $\tau_{\Delta} = 1.5$, and $\tau_{\mathrm{rel}} = 0.16$.

\paragraph{UniComp.}
We port the official UniComp implementation to Qwen3.5-0.8B.
It is training-free and does not see the question.
Frame-group fusion merges similar temporal groups using threshold $U_f$, token allocation spreads a budget of 64 tokens across the groups, and uniqueness-based selection keeps non-redundant tokens using the keys of the last vision block pooled over each $2 \times 2$ merge window with threshold $U_c$.
A short video can keep fewer than 64 tokens.
With the paper's $U_c = 0.2$, far fewer tokens count as non-redundant on Qwen3.5-0.8B keys than the paper reports, so we lower it to $U_c = 0.07$ to match the paper's selection ratio.
We set $U_f = 0.02$ by a sweep on held-out multiple-choice questions from the training splits of LLaVA-Video-178K, NExT-QA, and Perception Test.

\paragraph{Captioner (9B $\rightarrow$ 0.8B).}
Qwen3.5-9B captions each video once from the same 128-frame visual budget as the other baselines, with a question-agnostic prompt that asks for the setting, people and objects, what happens and in what order, and any on-screen text, capped at 1{,}024 completion tokens.
Qwen3.5-0.8B then answers each question from that caption alone, with no video tokens and no LoRA.
Both models are served with vLLM at temperature 0 with thinking disabled.

\paragraph{Compressor (9B $\rightarrow$ 0.8B).}
The compiler is Qwen3.5-9B with a LoRA of rank 16 and scale 32 on the same module types as VaW, and it reads the video followed by 512 readout tokens.
Every eighth last-layer readout state, 64 in total, passes through an MLP consisting of LayerNorm, a linear layer of width 1024, a GELU activation, and a zero-initialized linear projection to width $d$.
Each output is added to the embedding of one of 64 placeholder tokens in the interpreter's first user turn, and no LoRA is generated.
The model is initialized from an earlier weight-generation pre-training run with the same compiler and a different decoder, and fine-tuned on our fine-tuning mix.
We evaluate the checkpoint at step 14,000 of its 14,410-step epoch.

\paragraph{Doc2LoRA (layer-wise weight generation).}
The compiler and the interpreter are both Qwen3.5-0.8B, so compiler layer $j$ and interpreter layer $j$ correspond one to one.
For each layer $j$, 64 learned latents attend only to the readout states of compiler layer $j$ through 6 blocks of width 1024 with 8 heads, whose weights are shared across layers.
Learned queries, one for each combination of module type, rank index, and LoRA factor, then read the latents of that layer, and per-type linear maps turn the query states into the rows of $A$ and $B$.
This gives a LoRA of rank 64 and scale 128 at the same 150 sites as VaW.
The compiler uses a LoRA of rank 32, and no prefix tokens are generated.
Compared with the released Doc-to-LoRA configuration, which generates rank-8 LoRA for the MLP down projections only, this version uses VaW's readout tokens and LoRA sites at four times VaW's rank.

\clearpage
\section{Ablation Details}
\label{app:ablation}

\Cref{tab:ablation_full} lists the per-benchmark results behind \Cref{fig:scaling}.
Rows within a group of \Cref{tab:ablation}, and points along a line of \Cref{fig:scaling}, differ only in the setting they name.
Across groups, the 9B runs differ in a cap on the gain of the backward pass through each compiler layer.
The channel group of \Cref{tab:ablation}, \Cref{fig:scaling}b, and the $P = 64$ line of \Cref{fig:scaling}c use a cap of 1.25 with a decoder learning rate of $1 \times 10^{-4}$.
The other groups of \Cref{tab:ablation} and the $P = 0$ line of \Cref{fig:scaling}c use a cap of 1.1 with a decoder learning rate of $5 \times 10^{-5}$.
\Cref{fig:scaling}a uses no backward gain cap, a decoder learning rate of $1 \times 10^{-4}$, and a compiler learning rate of $5 \times 10^{-5}$.
The 0.8B line of \Cref{fig:scaling}c uses the 0.8B compiler recipe.
Leaving out the tokens at test time removes the prefix, as in the first term of \cref{eq:dual}.
The 4B interpreter trains with smaller micro-batches of at most 24,576 tokens and 8 rows, which comes to 24,705 steps for one epoch.

\begin{table}[h]
  \caption{Per-benchmark results behind \Cref{fig:scaling} (accuracy, \%).}
  \label{tab:ablation_full}
  \centering
  \small
  \setlength{\tabcolsep}{4.5pt}
  \begin{tabular}{lccccccc}
    \toprule
    Setting & EgoSchema & NExT-QA & LongVideoBench & Video-MME & MLVU & MVBench & Avg. \\
    \midrule
    \multicolumn{8}{l}{\textit{(a) Compiler size ($\rightarrow$ 0.8B, $P = 64$, $r = 64$)}} \\
    0.8B & 60.2 & 71.6 & 45.9 & 50.9 & 60.7 & 57.4 & 57.8 \\
    2B & 65.6 & 74.3 & 46.1 & 54.1 & 61.7 & 59.0 & 60.1 \\
    4B & 65.0 & 75.0 & 46.9 & 53.2 & 61.1 & 60.8 & 60.3 \\
    9B & 66.8 & 75.7 & 47.8 & 53.9 & 61.7 & 60.9 & 61.1 \\
    \midrule
    \multicolumn{8}{l}{\textit{(b) Interpreter size (9B $\rightarrow$, $P = 64$, $r = 64$)}} \\
    VaW, 0.8B & 69.6 & 76.0 & 48.3 & 55.1 & 62.2 & 61.7 & 62.2 \\
    In-context, 2B & 55.2 & 75.9 & 55.6 & 59.2 & 66.4 & 61.3 & 62.3 \\
    VaW, 2B & 70.2 & 79.8 & 51.2 & 59.6 & 67.2 & 65.5 & 65.6 \\
    In-context, 4B & 69.6 & 83.1 & 60.8 & 68.1 & 72.3 & 68.8 & 70.5 \\
    VaW, 4B & 76.6 & 83.5 & 54.5 & 61.8 & 69.9 & 68.9 & 69.2 \\
    \midrule
    \multicolumn{8}{l}{\textit{(c) LoRA rank}} \\
    9B $\rightarrow$ 0.8B, $r = 16$ (20.6 MB) & 66.4 & 75.7 & 45.8 & 54.9 & 61.2 & 62.1 & 61.0 \\
    9B $\rightarrow$ 0.8B, $r = 32$ (41.0 MB) & 67.0 & 75.9 & 48.0 & 54.1 & 61.2 & 61.7 & 61.3 \\
    9B $\rightarrow$ 0.8B, $r = 64$ (81.9 MB) & 69.6 & 76.0 & 48.3 & 55.1 & 62.2 & 61.7 & 62.2 \\
    0.8B $\rightarrow$ 0.8B, $r = 16$ (20.6 MB) & 61.8 & 71.5 & 47.1 & 52.2 & 60.2 & 57.5 & 58.4 \\
    0.8B $\rightarrow$ 0.8B, $r = 32$ (41.0 MB) & 61.4 & 71.3 & 44.7 & 51.0 & 59.4 & 57.8 & 57.6 \\
    0.8B $\rightarrow$ 0.8B, $r = 64$ (81.9 MB) & 62.0 & 71.5 & 45.0 & 52.3 & 58.4 & 57.3 & 57.8 \\
    9B $\rightarrow$ 0.8B, $P = 0$, $r = 16$ (20.4 MB) & 63.4 & 68.8 & 41.5 & 48.2 & 56.7 & 56.0 & 55.8 \\
    9B $\rightarrow$ 0.8B, $P = 0$, $r = 64$ (81.8 MB) & 58.4 & 69.4 & 43.3 & 47.6 & 54.4 & 55.6 & 54.8 \\
    \bottomrule
  \end{tabular}
\end{table}

\section{Memory}
\label{app:compute}

\Cref{tab:compute} compares the memory of a compiled video with the interpreter's KV cache on 128-frame Video-MME videos, which have a median length of 8,307 visual tokens in bf16.
For in-context reading, Qwen3.5-0.8B stores the KV cache and linear-attention state of all 8,307 video tokens, taking 112.4 MB and growing with the number of frames.
For VaW at $r = 16$, the stored state consists of 20.4 MB for the generated LoRA across all 150 adapted projections and 0.2 MB for the 64 prefix token embeddings, and its size stays fixed regardless of video length or compiler size.

\begin{table}[h]
  \caption{Per-video memory with the video kept across questions (bf16, 128 frames / 8,307 visual tokens).}
  \label{tab:compute}
  \centering
  \small
  \begin{tabular}{lcccc}
    \toprule
    Method & Rank $r$ & \#Tokens & Scales with frames? & Stored (MB) \\
    \midrule
    In-context (0.8B KV cache)       & -- & 8,307 & Yes & 112.4 \\
    VaW ($P = 0$)                    & 16 & 0     & No  & 20.4 \\
    VaW ($P = 64$, default)          & 16 & 64    & No  & 20.6 \\
    VaW ($P = 64$)                   & 32 & 64    & No  & 41.0 \\
    VaW ($P = 64$)                   & 64 & 64    & No  & 81.9 \\
    \bottomrule
  \end{tabular}
\end{table}

\clearpage
\section{Video-MME Breakdown}
\label{app:video_mme}

\begin{table}[t]
  \caption{Video-MME accuracy (\%) of VaW with weights only ($P = 0$) and with weights and tokens ($P = 64$), on selected task types and on each video length. $\Delta$ is the gain from the prefix tokens. All task types, with confidence intervals, are in \Cref{tab:video_mme_breakdown}.}
  \label{tab:video_mme_gain}
  \centering
  \small
  \setlength{\tabcolsep}{2.5pt}
  \begin{tabular}{@{}lcccccc@{}}
    \toprule
    & \multicolumn{3}{c}{0.8B $\rightarrow$ 0.8B} & \multicolumn{3}{c}{9B $\rightarrow$ 0.8B} \\
    \cmidrule(lr){2-4} \cmidrule(lr){5-7}
    Category & $P{=}0$ & $P{=}64$ & $\Delta$ & $P{=}0$ & $P{=}64$ & $\Delta$ \\
    \midrule
    \multicolumn{7}{l}{\textit{Task type}} \\
    Attribute perception & 52.3 & 59.9 & +7.6 & 58.6 & 67.6 & +9.0 \\
    Object recognition & 42.4 & 59.9 & +17.5 & 47.5 & 61.3 & +13.8 \\
    OCR & 43.2 & 64.7 & +21.5 & 38.8 & 59.7 & +20.9 \\
    Information synopsis & 62.5 & 67.2 & +4.7 & 65.3 & 70.0 & +4.7 \\
    \midrule
    \multicolumn{7}{l}{\textit{Video length}} \\
    Short & 51.7 & 64.0 & +12.3 & 56.4 & 67.1 & +10.7 \\
    Medium & 43.3 & 51.2 & +7.9 & 46.0 & 51.7 & +5.7 \\
    Long & 38.0 & 41.4 & +3.4 & 42.2 & 46.0 & +3.8 \\
    \bottomrule
  \end{tabular}
\end{table}

\Cref{tab:video_mme_breakdown} lists the Video-MME accuracy of VaW with $P = 0$ and with $P = 64$ for every task type and video length, with 95\% confidence intervals for the gains in \Cref{tab:video_mme_gain}.
The runs are the $r = 16$ checkpoints of \Cref{tab:main_results}.
For both compilers, the gains on object recognition, OCR, and short and medium videos exclude zero.
Counting, temporal reasoning, and spatial perception stay close to or below the blind score, so their differences carry little information.
Task types with fewer than 100 questions have wide intervals.

\begin{table*}[ht]
  \caption{Video-MME accuracy (\%) by task type and video length, for VaW with weights only ($P = 0$) and with weights and tokens ($P = 64$). $\Delta$ is the gain from the prefix tokens, and CI is its 95\% confidence interval from a normal approximation over questions. $\Delta$ is in \textbf{bold} when the interval excludes zero. Blind is the interpreter without the video. Task types are ordered by their average $\Delta$, and $^{\dagger}$ marks those with fewer than 100 questions. Runs match \Cref{tab:main_results}: $r = 16$.}
  \label{tab:video_mme_breakdown}
  \centering
  \small
  \setlength{\tabcolsep}{4pt}
  \begin{tabular}{lrrcccccccc}
    \toprule
    & & & \multicolumn{4}{c}{VaW (0.8B $\rightarrow$ 0.8B)} & \multicolumn{4}{c}{VaW (9B $\rightarrow$ 0.8B)} \\
    \cmidrule(lr){4-7} \cmidrule(lr){8-11}
    Category & \#Q & Blind & $P = 0$ & $P = 64$ & $\Delta$ & CI & $P = 0$ & $P = 64$ & $\Delta$ & CI \\
    \midrule
    \multicolumn{11}{l}{\textit{Task type}} \\
    OCR & 139 & 25.9 & 43.2 & 64.7 & \textbf{+21.5} & {\footnotesize [+10.1, +32.9]} & 38.8 & 59.7 & \textbf{+20.9} & {\footnotesize [+9.4, +32.4]} \\
    Object recognition & 354 & 28.5 & 42.4 & 59.9 & \textbf{+17.5} & {\footnotesize [+10.2, +24.8]} & 47.5 & 61.3 & \textbf{+13.8} & {\footnotesize [+6.5, +21.1]} \\
    Temporal perception\smash{$^{\dagger}$} & 55 & 40.0 & 40.0 & 63.6 & \textbf{+23.6} & {\footnotesize [+5.5, +41.7]} & 60.0 & 56.4 & $-$3.6 & {\footnotesize [$-$22.0, +14.8]} \\
    Attribute perception & 222 & 31.1 & 52.3 & 59.9 & +7.6 & {\footnotesize [$-$1.6, +16.8]} & 58.6 & 67.6 & \textbf{+9.0} & {\footnotesize [+0.1, +17.9]} \\
    Spatial reasoning\smash{$^{\dagger}$} & 56 & 33.9 & 62.5 & 62.5 & +0.0 & {\footnotesize [$-$17.9, +17.9]} & 55.4 & 69.6 & +14.2 & {\footnotesize [$-$3.5, +31.9]} \\
    Object reasoning & 454 & 30.8 & 36.8 & 45.4 & \textbf{+8.6} & {\footnotesize [+2.2, +15.0]} & 41.6 & 46.9 & +5.3 & {\footnotesize [$-$1.2, +11.8]} \\
    Information synopsis & 323 & 41.5 & 62.5 & 67.2 & +4.7 & {\footnotesize [$-$2.7, +12.1]} & 65.3 & 70.0 & +4.7 & {\footnotesize [$-$2.5, +11.9]} \\
    Action recognition & 313 & 28.4 & 47.9 & 53.0 & +5.1 & {\footnotesize [$-$2.7, +12.9]} & 51.4 & 55.0 & +3.6 & {\footnotesize [$-$4.2, +11.4]} \\
    Action reasoning & 285 & 31.6 & 41.8 & 46.0 & +4.2 & {\footnotesize [$-$3.9, +12.3]} & 44.6 & 49.1 & +4.5 & {\footnotesize [$-$3.7, +12.7]} \\
    Spatial perception\smash{$^{\dagger}$} & 54 & 53.7 & 44.4 & 51.9 & +7.5 & {\footnotesize [$-$11.3, +26.3]} & 59.3 & 59.3 & +0.0 & {\footnotesize [$-$18.5, +18.5]} \\
    Counting & 268 & 32.5 & 36.9 & 39.9 & +3.0 & {\footnotesize [$-$5.2, +11.2]} & 40.3 & 42.2 & +1.9 & {\footnotesize [$-$6.4, +10.2]} \\
    Temporal reasoning & 177 & 37.9 & 29.9 & 28.2 & $-$1.7 & {\footnotesize [$-$11.2, +7.8]} & 32.8 & 37.9 & +5.1 & {\footnotesize [$-$4.8, +15.0]} \\
    \midrule
    \multicolumn{11}{l}{\textit{Video length}} \\
    Short & 900 & 32.9 & 51.7 & 64.0 & \textbf{+12.3} & {\footnotesize [+7.8, +16.8]} & 56.4 & 67.1 & \textbf{+10.7} & {\footnotesize [+6.2, +15.2]} \\
    Medium & 900 & 31.4 & 43.3 & 51.2 & \textbf{+7.9} & {\footnotesize [+3.3, +12.5]} & 46.0 & 51.7 & \textbf{+5.7} & {\footnotesize [+1.1, +10.3]} \\
    Long & 900 & 33.8 & 38.0 & 41.4 & +3.4 & {\footnotesize [$-$1.1, +7.9]} & 42.2 & 46.0 & +3.8 & {\footnotesize [$-$0.8, +8.4]} \\
    \midrule
    \rowcolor{gray!12}All & 2,700 & 32.7 & 44.3 & 52.2 & \textbf{+7.9} & {\footnotesize [+5.2, +10.6]} & 48.2 & 54.9 & \textbf{+6.7} & {\footnotesize [+4.0, +9.4]} \\
    \bottomrule
  \end{tabular}
\end{table*}

\section{Weight Analysis Details}
\label{app:weight_analysis}

Here we give the formal definitions for the two analyses in \Cref{sec:weight_analysis}, evaluated on VaW (9B $\rightarrow$ 0.8B) with $P = 64$ and $P = 0$ across the 900 Video-MME videos.

\paragraph{Compiler-to-interpreter layer attribution.}
To isolate the video-specific contribution of compiler layer $\ell$ in \Cref{fig:layer_corr}, we replace its readout states with their average over all other videos at the same relative positions, yielding $H^{(\ell)}$.
Replacing all compiler layers at once gives the video-agnostic baseline $\bar{H}$.
We then measure how much the generated LoRA of interpreter layer $j$ changes relative to removing all video-specific information:
\begin{equation}
  e_{\ell,j} = \mathbb{E}_{V}\!\left[\frac{\|\Delta W_j(H) - \Delta W_j(H^{(\ell)})\|}{\|\Delta W_j(H) - \Delta W_j(\bar{H})\|}\right],
  \label{eq:layer_share}
\end{equation}
where $\Delta W_j$ stacks the generated LoRA updates across all adapted sites in interpreter layer $j$.
Each row of \Cref{fig:layer_corr} reports $e_{\ell,j}$ normalized to sum to one over $\ell \in \{1, \dots, L\}$, and the ``emb'' row measures the prefix token embeddings $\mathbf{P}$ in the same way.

\paragraph{Per-module output change.}
To measure how strongly the generated LoRA changes each interpreter module in \Cref{fig:module_change}, we compare the output norm of the rank-$r$ update with that of the frozen base weight.
For site $s$ receiving input activations $\mathbf{u}_t$ at token $t$,
\begin{equation}
  r_s = \left(\frac{\sum_{t} \|\Delta W_s \mathbf{u}_t\|^2}{\sum_{t} \|W_s \mathbf{u}_t\|^2}\right)^{1/2},
  \label{eq:module_change}
\end{equation}
where $t$ runs over the prompt tokens of the video's questions, and we report the median across the 900 Video-MME videos.

\end{document}